\pdfoutput=1
\documentclass[11pt]{article}

\usepackage[final]{acl}

\usepackage{times}
\usepackage{latexsym}

\usepackage[T1]{fontenc}
\usepackage[utf8]{inputenc}

\usepackage{microtype}

\usepackage{inconsolata}

\usepackage{graphicx}

\usepackage{amsmath, amsfonts}
\newcommand{\model}{RelAdapter}
\newcommand{\stitle}[1]{\vspace{1mm}\noindent\textbf{#1}}
\usepackage{algorithm} 
\usepackage{algpseudocode} 
\usepackage{booktabs}
\usepackage{subcaption}
\usepackage{multirow}

\usepackage{pifont}
\usepackage{tabu}

\usepackage{booktabs} % For better looking tables

\title{Context-Aware Adapter Tuning for Few-Shot Relation Learning in Knowledge Graphs}

\author{Ran Liu$^{1}$, Zhongzhou Liu$^1$, Xiaoli Li$^{2,3}$ \and Yuan Fang$^{1}$\\
        $^1$School of Computing and Information Systems, Singapore Management University, Singapore\\
        $^2$Institute for Infocomm Research, A*STAR, Singapore\\ $^3$A*STAR Centre for Frontier AI Research, Singapore\\
        \{ran.liu.2020, zzliu.2020\}@phdcs.smu.edu.sg,  xlli@i2r.a-star.edu.sg, yfang@smu.edu.sg}

\begin{document}

\maketitle
\begin{abstract}
  Knowledge graphs (KGs) are instrumental in various real-world applications, yet they often suffer from incompleteness due to missing relations. To predict instances for novel relations with limited training examples, few-shot relation learning approaches have emerged, utilizing techniques such as meta-learning. However, the assumption is that novel relations in meta-testing and base relations in meta-training are independently and identically distributed, which may not hold in practice. To address the limitation, we propose \model, a context-aware adapter for few-shot relation learning in KGs designed to enhance the adaptation process in meta-learning. First, \model\ is equipped with a lightweight adapter module that facilitates relation-specific, tunable adaptation of meta-knowledge in a parameter-efficient manner. Second, \model\ is enriched with contextual information about the target relation, enabling enhanced adaptation to each distinct relation. Extensive experiments on three benchmark KGs validate the superiority of \model\ over state-of-the-art methods. 
  
  % Knowledge graphs (KGs) are instrumental in various real-world applications, yet they often suffer from incompleteness due to missing relations. To predict instances for novel relations with limited training examples, few-shot relation learning approaches have emerged, utilizing techniques such as meta-learning. By using meta-learning, it is assumed that novel relations in meta-testing and base relations in meta-training are independently and identically distributed, which may not hold for novel relations in practice. To address the limitation, we propose \model, a context-aware adapter for few-shot relation learning in KGs designed to enhance the adaptation process in meta-learning. First, \model\ is equipped with a lightweight adapter module that facilitates relation-specific, tunable adaptation of meta-knowledge in a parameter-efficient manner. Second, \model\ is enriched with contextual information about the target relation, enabling enhanced adaptation to each distinct relation. Extensive experiments on three benchmark KGs validate the superiority of \model\ over state-of-the-art methods.
\end{abstract}

\section{Introduction}

Knowledge graphs (KGs) \cite{bollacker2008freebase,suchanek2007yago,vrandevcic2014wikidata} have been widely adopted to describe real-world facts using triplets in the form of (head entity, relation, tail entity). 
%Given a vast collection of triplets, various entities can be interconnected to form a large-scale KG \cite{bollacker2008freebase,suchanek2007yago,vrandevcic2014wikidata}, which can provide powerful knowledge inference capabilities to enhance numerous intelligent applications, including question answering \cite{shin2019predicate,wu2016ask}, object detection \cite{fang2017object} and recommendation \cite{cao2019unifying,du2021diversity,isufi2021accuracy}. 
%Despite such importance, 
However, curating and maintaining  all the possible ground-truth triplets is impossible, and various approaches for knowledge graph completion \cite{bordes2013translating,yang2014embedding,trouillon2016ComplEx,sun2019rotate} have been proposed to discover missing facts. Many of these methods adopt a supervised learning paradigm, 
which require abundant training data for each relation.
%and their performance depend on the availability of abundant training data for each relation.
In real-world settings, novel and emerging relations, along with many relations in the long tail, are associated with very few instances \cite{xiong2018one}, limiting their performance.
%of conventional supervised approaches on these relations.

Subsequently, \emph{few-shot relation learning} (FSRL) on KGs has emerged to handle novel relations with only a few known instances. 
An established line of work \cite{chen2019meta,niu2021relational} employs \emph{meta-learning}, most notably Model-Agnostic Meta-Learning (MAML) \cite{finn2017model}. MAML aims to learn a prior from a series of meta-training tasks, which can be rapidly adapted to downstream meta-testing tasks. The meta-training tasks are specifically constructed in a few-shot setup to mimic downstream tasks. 
%Given its superior performance on few-shot tasks, meta-learning has also been extended to FSRL  \cite{chen2019meta,niu2021relational}. 
%are designed to learn a prior knowledge during upstream training which can be easily fined-tuned under few-shot setting to adapt to downstream relation tasks. 
In the context of FSRL, each task contains only a few training instances of a single relation, and the objective is to predict more instances for a novel task (i.e., relation\footnote{We use the terms ``task'' and ``relation'' interchangeably.}) not seen in meta-training. For example, MetaR \cite{chen2019meta} aims to learn a relation-specific prior (also called meta-knowledge) during the meta-training stage, using a series of meta-training tasks constructed from a set of base relations with abundant instances. Subsequently, the meta-knowledge is leveraged for rapid adaptation, through a lightweight fine-tuning step, for few-shot predictions on novel relations in meta-testing.

\begin{figure}[t]
    \centering
    \hspace{-3mm}    \subfloat[\centering WIKI]{{\includegraphics[width=2.63cm]{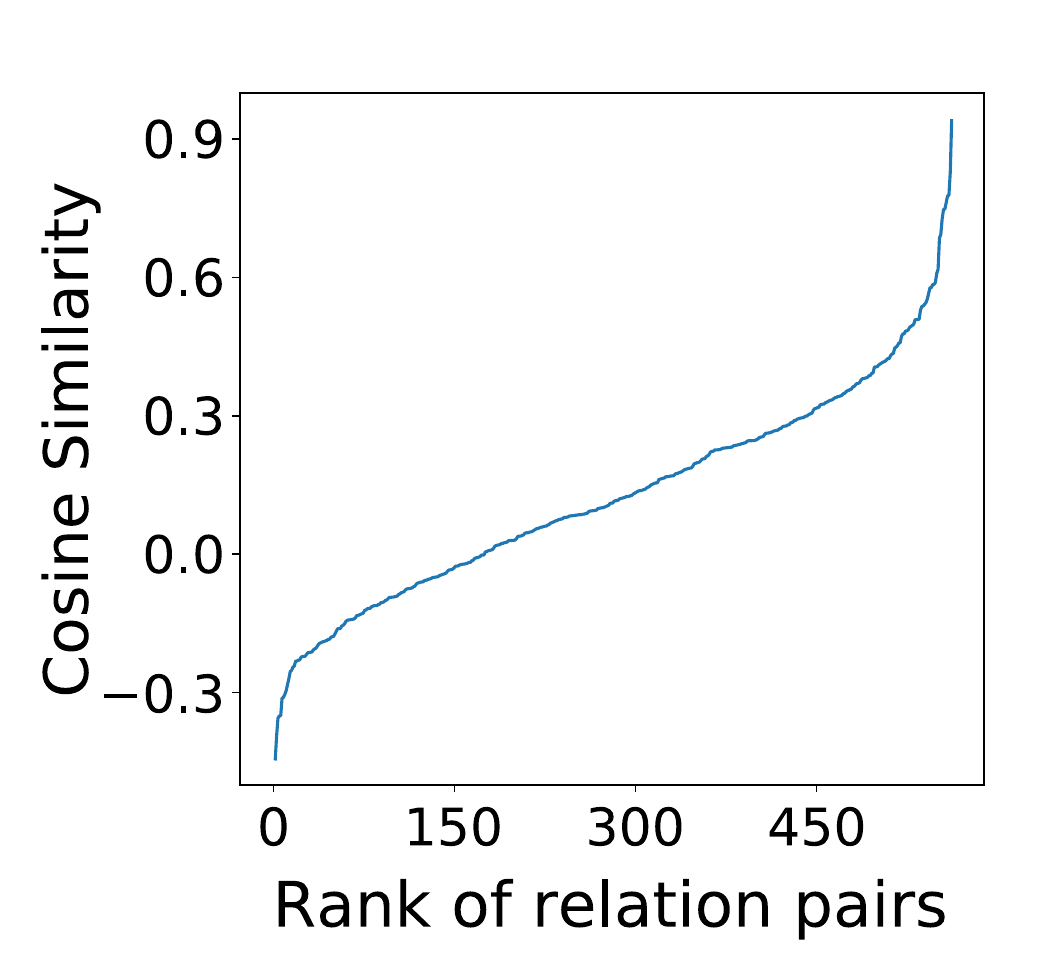} }}
    \hspace{-3mm}
    \subfloat[\centering FB15K-237]{{\includegraphics[width=2.63cm]{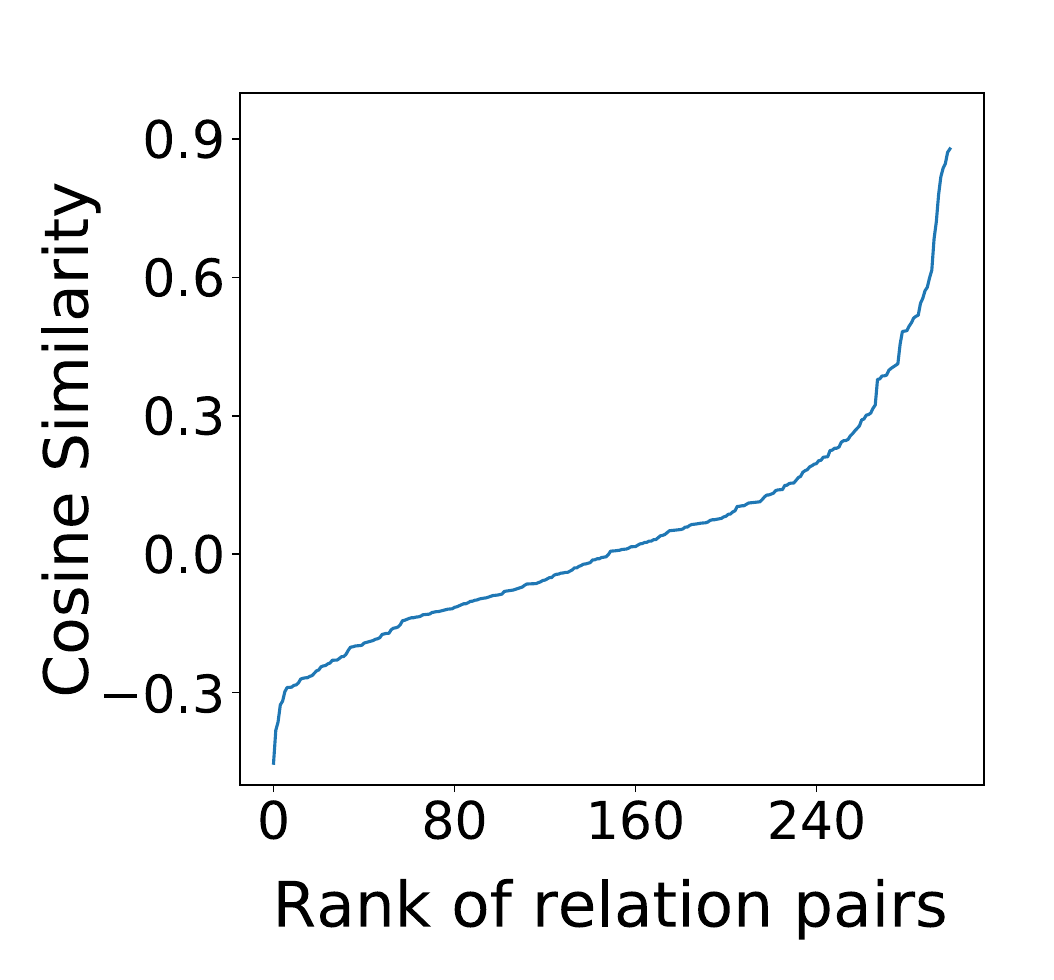} }}
    \hspace{-3mm}
    \subfloat[\centering UMLS]{{\includegraphics[width=2.63cm]{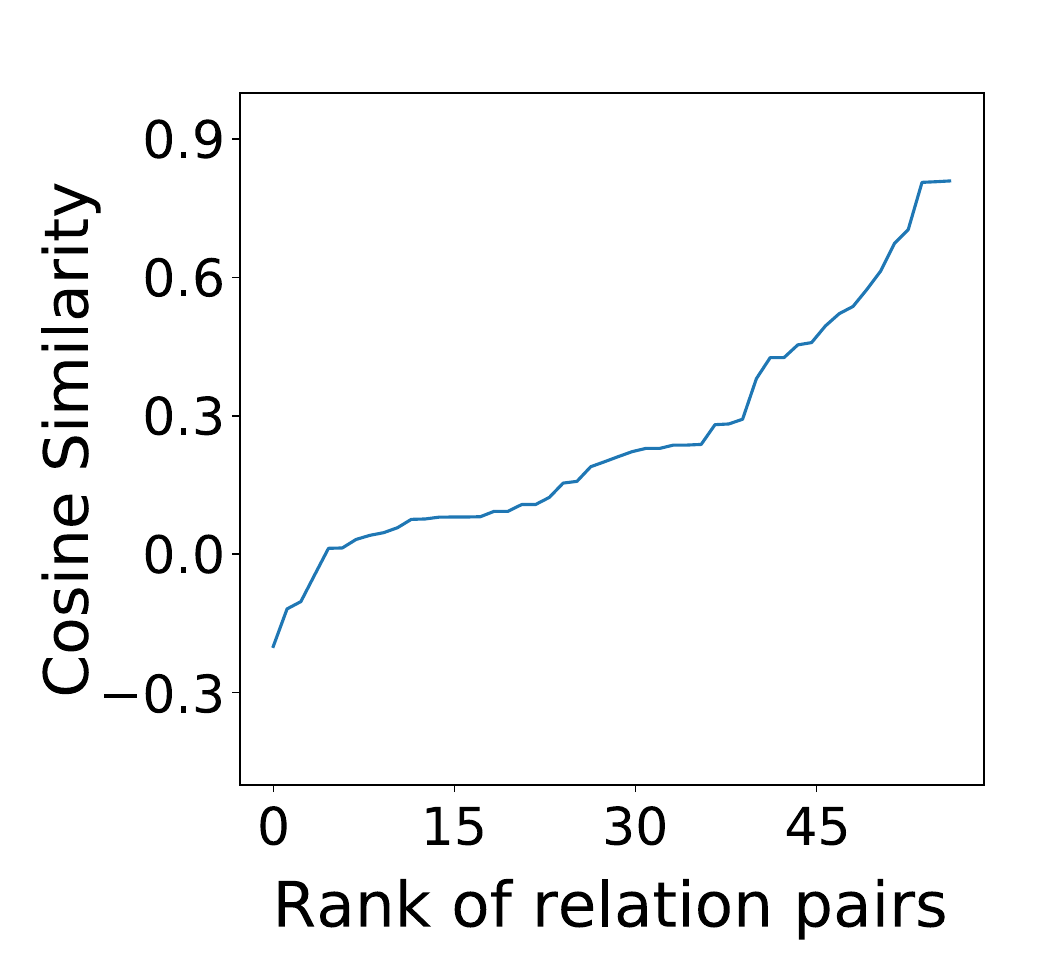} }}
    \hspace{-3mm}\vspace{-2mm}
    \caption{Pairwise cosine similarity of relations.}
    \label{fig:pilot}
\end{figure}

\stitle{Limitation of prior work.} %Despite their favorable performance compared to conventional supervised approaches, FSRL methods based on meta-learning are not without their drawback. 
The major limitation of FSRL methods based on meta-learning lies in the assumption that the meta-training and meta-testing tasks are independently and identically distributed (i.i.d.). %Essentially, to enable effective adaptation, the novel relations in meta-testing shall be drawn from the same task distribution as the meta-training tasks on the base relations. 
However, different relations may diverge significantly in their underlying distributions, thereby weakening the i.i.d.~task assumption. To investigate this hypothesis, we randomly sample a large number of relation pairs from standard benchmark datasets, namely, WIKI, FB15K-237 and UMLS\footnote{Refer to Sect.~\ref{sec:expt} for dataset details.}. We perform a mean pooling across all entities within each relation task to derive an average embedding as the task representation. For every pair of relations, we plot their cosine similarity in Fig.~\ref{fig:pilot}. The results reveal a wide variance in the similarities between relations, suggesting that a uniform adaptation process may not suffice for all relations. In particular, for out-of-distribution relations, performance degradation is generally anticipated \cite{radstok2021knowledge,li2022ood}.
%, unless the distribution shift between base and novel relations are explicitly addressed. 

\stitle{Challenges and insights.}
%
%
%The second drawback of Fewshot relational learning (FSRL) models is the (2) Lack of graph structural information in Data Augmentation. Existing methods \cite{xiong2018one,zhang2020few,sheng2020adaptive,chen2019meta,niu2021relational,wu2023hierarchical} focus on different model designs to extract knowledge during meta-training for downstream relation predictions. Such methods do not fundamentally solve the issue of data sparsity which led to the rise of few-shot learning methodologies.  Subsequently, data augmentation models such as \cite{pei2023few,wu2023hierarchical} introduced inter-class and intra-class augmentation to improve the FSRL performances. However, these models do not introduce existing graph structural information during the data augmentation process.
%
Specifically, we identify two open challenges, at the model and data levels, towards more universal downstream adaptation. % to a  more universal range of relations. % during meta-testing. 
At the \emph{model level}, the first challenge (\textbf{C1}) is how to design a relation-specific adaptation module that can be tailored to each relation, while still leveraging the meta-knowledge learned from meta-training. The goal is to enable a more flexible adaptation process that allows for a relation-specific balance between the global prior and local task input.
%to better  effective strategy that adapts the meta prior learnt to fit downstream out-of-distribution relation tasks. The solution is non-trivial as the distributions of each relation task may be discretely different and the design of the model needs to cater to each individual task. 
At the \emph{data level}, the second challenge (\textbf{C2}) is how to augment few-shot relation instances during meta-testing in an unsupervised manner, thereby further enhancing adaptation to novel relations. 

%apply data augmentation by incorporating additional graph structural information to downstream few-shot relation tasks. As \cite{xiong2018one} suggested that explicitly encoding graph local structure can benefit few-shot relation prediction, we believe this could also benefit data augmentation in FSRL.

% Task distribution differences are not new in the paradigm of natural language processing (NLP). With the prevalence of large pre-trained language models and many downstream tasks, it is inefficient to tune the whole model for each task. As such, some researchers turn to adapters \cite{houlsby2019parameter,pfeiffer2020adapterfusion} for transfer learning in NLP. In a similar spirit for graphs, AdapterGNN \cite{li2023adaptergnn} incorporates task-specific knowledge to tune a dual adapter module that bridges the gap between transformer based models and graph neural network. However, empirical studies \cite{gui2023g} have shown that directly transferring existing parameter-efficient fine-tuning (PEFT) to graph-based tasks may not generate favorable performances as a result of feature distribution shift. Therefore,  G-Adaptor \cite{gui2023g} introduces graph structure knowledge as a training constraint which guides the updating of adaptor parameters. In view of the above, our research take inspirations across other fields and aim to incorporate the use of adapter into the field of meta-learning. Idential to G-Adaptor \cite{gui2023g}, \model\ leverages on contextual graph structural information to aid the training of the adaptor.

To address these challenges, we introduce a context-aware adapter framework, called \model, for few-shot relation learning. To overcome the model-level challenge (C1), we draw inspiration from parameter-efficient fine-tuning \cite{houlsby2019parameter,hu2021lora,li2021prefix,brown2020language}.
%, a concept originated in natural language processing. 
Specifically, we propose to integrate a lightweight \emph{adapter} module into the meta-learning framework. The adapter module enables a relation-specific, tunable adaptation of the global prior to suit the local task in the meta-testing stage. %Meanwhile, the adapter constitutes a small neural network that is amenable to FSRL, with a lower propensity to overfit.
%Firstly, motivated by AdapterGNN \cite{li2023adaptergnn} that suggests the use of adapters to bridge the gap between transformer based models and graph neural network, we propose the addition of an adapter that adjusts upstream prior learnt during training to downstream relation tasks despite the differences in task distributions. Secondly, taking inspirations from G-Adapter \cite{gui2023g} which leverages on contextual graph structure to aid the training of the adapter, we incorporate existing knowledge (graph structures) from the pre-training to augment the entity embeddings of downstream relation tasks.
For the data-level challenge (C2), we propose to inject additional \emph{contextual information} about the target relation into meta-testing. The contextual information is extracted based on existing KG structures without requiring any extra annotation, serving as a form of data augmentation to enrich the few-shot relation instances.  This strategy endows the adapter with more relation-specific contexts, and the context-aware adapter can enhance relation-specific adaptation.
%The context-aware adapter enables more precise relation-specific customization, leading to a granular adaptation to each distinct relation.    

%apply data augmentation by incorporating additional graph structural information to downstream few-shot relation tasks. As \cite{xiong2018one} suggested that explicitly encoding graph local structure can benefit few-shot relation prediction, we believe this could also benefit data augmentation in FSRL.

\stitle{Contributions.}
In summary, we make the following contributions. (1) We observe the i.i.d.~limitation of prior work in FSRL, and first support it with an empirical analysis. (2) We propose \model, a context-aware adapter framework for FSRL. Specifically, we design a lightweight adapter module that leverages contextual information, enabling a relation-specific, tunable, and fine-grained adaptation for each novel relation in meta-testing. (3) We conduct extensive experiments on three benchmark datasets, and demonstrate the superiority of our proposed \model.

\section{Related Work}

%In this section, we review literature from three main categories: supervised relation learning, few-shot relation learning, and adapters.

\stitle{Supervised relation learning.}
Knowledge graph embedding aims to transform entities and relations into a low-dimensional continuous vector space while preserving their semantic meaning. Conventional knowledge graph completion models can generally be classified into three main categories: (1) Translation-based methods, such as TransE \cite{bordes2013translating}, TransH \cite{wang2014knowledge}, and TransD \cite{ji2015knowledge}, which are additive models that use distance-based constraint to optimize entity and relations embedding. 
%For instance, TransE \cite{bordes2013translating} executes triplet level translations between head and tail entities in knowledge graphs to model the connections to satisfy the translational constraint$\|\vec{e}_h+\vec{e}_r - \vec{e}_t\|$. TransH \cite{wang2014knowledge} builds on TransE by differentiating entity embedding on relation-specific hyper-planes, enriching entities with various representations. TransD \cite{ji2015knowledge} further improves on past models by mapping entities and relations as objects living in different vector spaces with different entity-relation-specific projections. Despite the extensive research, such models are often associated with having low expressive power and its inability to capture semantic information. 
(2) Semantic matching-based methods, such as DistMult \cite{yang2014embedding} and ComplEx \cite{trouillon2016ComplEx}, which are multiplicative models that exploit the interaction between entity and relation vectors.
%
%. Such models include DistMult \cite{yang2014embedding} that applies a bi-linear interaction between entity and relation vector as an element of diagonal matrix. ComplEx \cite{trouillon2016ComplEx} builds upon DistMult by tapping the ComplEx vector space for embedding entities and relations, which improved the expressive power for knowledge graph embedding. However, similar to translation-based methods, these models only account for individual triplet but neglects the intrinsic knowledge graph structure which contains deeper semantics. 
(3) Graph-based models, include graph neural networks such as GCN \cite{kipf2016semi} and RGCN \cite{schlichtkrull2018modeling}, which considers higher-order structures in KGs.
%Such models dwells deep into the interconnections of triplets and higher order information in knowledge graphs. 
%In addition, the use of convolution neural networks have proven to produce better embedding. 
%Despite having achieved state-of-art performances for datasets with abundant training instances, 
%
However, these supervised approaches rely on a large amount of training data and are not well-suited for few-shot relation learning. 

\stitle{Few-shot relation learning.}
%In reality, few-shot knowledge graph completion models are more suitable for real-world datasets. These models are focused on making accurate relation predictions despite having limited training samples. 
To address one- or few-shot relation learning, many models have been proposed recently in two main categories: (1) Metric-based models that calculates a similarity score between support and query sets to learn the matching metrics. GMatching \cite{xiong2018one} %is the first work that is formulated for one-shot FSRL. It 
uses a one-hop neighbor encoder and a matching network, but assumes that all neighbors contribute equally. FSKGC \cite{zhang2020few} extends the setting to more shots and seeks to merge information learnt from multiple reference triplets with a fixed attention mechanism. FAAN \cite{sheng2020adaptive} introduces an relation-specific adaptive neighbor encoder for one-hop neighbors. (2) Optimization-based models aim to learn an initial meta prior that can be generalized to a new relation given few-shot examples. MetaR \cite{chen2019meta} makes predictions by transferring shared relation-specific meta information from the support set to the queries through a mean pooling of the support set. GANA \cite{niu2021relational} improves on previous models by having a gated and attentive neighbor aggregator to capture the most valuable contextual semantics of a few-shot relation. HiRe \cite{wu2023hierarchical} further considers triplet-level contextual information to generate relation meta.
%which is further constrained by pair-wise entities. 
Meanwhile, REFORM \cite{wang2021reform} seeks to alleviate the impact of potential errors, which can be prevalent in KGs and further exacerbated under few-shot settings.

Despite these efforts, they fail to explicitly account for out-of-distribution relations. NP-FKGC \cite{luo2023normalizing} offers a solution to the out-of-distribution problem by leveraging neural processes to model complex distributions which can fit both training and test data. Unfortunately, it cannot be applied directly to existing meta-learning frameworks, and is therefore unable to benefit from meta-learned prior.
Similarly, several other recent works \cite{liu2024few,meng2024sarf,huang2022few} for few-shot relation learning do not utilize meta-learning frameworks, and thus do not leverage a set of meta-training tasks to learn a prior, as done in our work. %Therefore, these methods cannot be compared to our method. In contrast, in our method and the baselines \cite{chen2019meta,niu2021relational, wu2023hierarchical,sheng2020adaptive} we compared to,  the meta-training tasks are utilized to learn the prior which ensures a fair and consistent comparison.

\stitle{Adapters.}
Parameter-efficient fine-tuning has gained traction in various domains \cite{houlsby2019parameter,li2021prefix,brown2020language,hu2021lora}. The adapter \cite{houlsby2019parameter} is a common parameter-efficient technique, which adds a sub-module to pre-trained language models to adapt them to downstream tasks. In the field of general graph learning, AdapterGNN \cite{li2023adaptergnn} employs adapters to bridge the gap between transformer-based models and graph neural networks. On the other hand, G-Adapter \cite{gui2023g} leverages  contextual graph structures as an inductive bias to aid the training of the adapter. 
Both approaches are designed for general graph learning, and cannot handle the problem of FSRL on KGs. 
Furthermore, our work focuses on integrating adapters into meta-learning frameworks to leverage meta-learned knowledge, which are missing in prior adapter designs for graphs.

\stitle{Contextual information.} Previous works have utilized contextual information in knowledge graphs. For example, \citet{oh2018knowledge} tap on the contexts from multi-hop neighborhoods, and \citet{tan2023kracl} apply an attention mechanism to aggregate neighboring contexts.  However, how contextual information can be utilized for few-shot relation learning, particularly within an adapter module, remains unclear. In contrast, our work introduces a context-aware adapter module to address the few-shot setting.

% In \model\, the contextual information empowers the adapter and enables more tailored adaptation to each distinct novel relation under few-shot settings.

\section{Preliminaries}\label{sec:prelim}
In this section, we introduce the problem of few-shot relation learning (FSRL) and the meta-learning framework for this problem.

\stitle{Problem formulation.}
A knowledge graph $\mathcal{G}=(\mathcal{V},\mathcal{R})$ comprises a set of triplets. Each triplet is represented by the form $(h,r,t)$ for some $h, t\in \mathcal{V}$ and $r\in \mathcal{R}$, where  $\mathcal{V}$ is the set of entities and $\mathcal{R}$ is the set of relations. 
% \lr{In addition, there exists a background graph $\mathcal{G}_{b}$ which takes the same form as $\mathcal{G}$ for all entities $h, t\in \mathcal{V}$. For the purpose of pre-training, the relations in $r\in \mathcal{R}_{b}$ are unseen from $r\in \mathcal{R}$ such that $\mathcal{R} \cap \mathcal{R}_{b} = \varnothing$.}

%entities $\mathcal{V}$, and a set of relations $\mathcal{R}$. defined by a set of triplets, where each triplet $\langle h,r,t\rangle$ for some $h, t\in \mathcal{V}$ and $r\in \mathcal{R}$, where $\mathcal{V}$ and $\mathcal{R}$ represent the entity set and relation set respectively. 

%\stitle{Few-shot Relation Learning (FSRL).} 
Consider a novel relation $r\notin \mathcal{R}$ w.r.t.~a knowledge graph $\mathcal{G}$. Further assume a support set $\mathcal{S}_r$ =  $\{(h_{i}, r, t_{i})\mid i=1,2,\ldots, K\}$ for some $h_{i}, t_{i} \in \mathcal{V}$ for the novel relation $r$. 
The goal is to predict the missing tail entities in a query set, $\mathcal{Q}_r$ = $\{(h_j, r, ?) \mid j=1,2,\ldots, \}$, w.r.t.~the given $h_j\in \mathcal{V}$ and $r$.

In the paper, for each triplet in $(h_{j}, r, ? ) \in \mathcal{Q}_r$, a type-constrained candidate set $\mathcal{C}_{h_{j},r}$ is provided and the objective is to rank the true tail entities highest among the candidates. Together, the support and query sets form a task $\mathcal{T}_r=(\mathcal{S}_r,\mathcal{Q}_r)$ for $r$.
The support set provides a few training instances, known as $K$-shot relation learning, where $|\mathcal{S}_r| = K$ is typically a small number. 

\stitle{Meta-learning framework.} 
%\label{sec:backbone}
Given the success of meta-learning in few-shot problems, we adopt MetaR \cite{chen2019meta}, a popular meta-learning-based approach for FSRL, as our learning framework.
% Owing to the success of MetaR model, in this paper, we adopt it as the backbone for $\model$. 
MetaR consists of two stages: \emph{meta-training} and \emph{meta-testing}, aiming to learn a prior $\Phi$ from the meta-training stage that can be adapted to the meta-testing stage.
On one hand, meta-training involves a set of seen relations $\mathcal{R}^\text{tr}$, and operates on their task data $\mathcal{D}^{\text{tr}}  =\{\mathcal{T}_r\mid r\in \mathcal{R}^\text{tr}\}$.
On the other hand, meta-testing involves a set of novel relations $\mathcal{R}^\text{te}$ such that $\mathcal{R}^\text{tr} \cap \mathcal{R}^\text{te}=\emptyset$, and operates on their task data   
$\mathcal{D}^{\text{te}}  =\{\mathcal{T}_r\mid r\in \mathcal{R}^\text{te}\}$. Note that the ground-truth tail entities are provided in the query sets of the meta-training tasks $\mathcal{D}^{\text{tr}}$, whereas the objective is to make predictions for the query sets of the meta-testing tasks $\mathcal{D}^{\text{te}}$.

% but are awaiting prediction in the meta-testing tasks $\mathcal{D}^{\text{te}}$.
%
%The training process is based on a set of tasks $\mathcal{T}_{\text{tr}}$ = \{$\mathcal{T}_{\text{r}}\}^{|\mathcal{R}_{\text{train}}|}_{r = 1}$ for $\text{r} \in \mathcal{R}_\text{train}$
% each $\mathcal{T}_{A}$ = $\{\mathcal{S}_{r}, \mathcal{Q}_{r}\}$ contains its own support/query set for relation $r\in \mathcal{R}$.
%and $|\mathcal{R}_{\text{train}}|$ denotes the number of training task relations in $\mathcal{T}_{\text{tr}}$. 
%The testing process is conducted on a new set of relation tasks
%$\mathcal{T}_{\text{te}}$  =\{$\mathcal{T}_{\text{r}}\}^{|\mathcal{R}_{\text{test}}|}_{r = 1}$ for $\text{r} \in \mathcal{R}_\text{test}$ and $|\mathcal{R}_{\text{test}}|$ denotes the number of testing task relations in $\mathcal{T}_{te}$. 
%The task relations in training and testing are both derived from the set of relations in  $\mathcal{G}$ for all $\mathcal{R}_{\text{train}},\mathcal{R}_{\text{test}} \in R$. However, the relations in  $\mathcal{T}_{\text{te}}$ are not seen in the  $\mathcal{T}_{\text{tr}}$ such that $\mathcal{R}_{\text{train}} \cap \mathcal{R}_{\text{test}} = \varnothing$. 

\stitle{Meta-training.} During meta-training, the model learns a prior $\Phi$, which serves as a good initialization to extract a shared \emph{relation meta}, $R_{\mathcal{T}_r}\in \mathbb{R}^d$, for each task $\mathcal{T}_r = (\mathcal{S}_r,\mathcal{Q}_r) \in \mathcal{D}^{\text{tr}}$. Specifically, the relation meta $R_{\mathcal{T}_r}$ is obtained through a mean pooling of all (head, tail) pairs in the support set of the task $\mathcal{T}_r$, as follows.
\begin{align}
    R_{\mathcal{T}_r}
  %  R_{\left(h_i, t_i\right)}
    =\mathtt{Mean}(\{\mathtt{RML}(f({h}),f({t});\Phi)\mid  \mathcal{S}_r\}),
    \label{eqn:relation-meta}
 \end{align}
 % \begin{align}
 %    R_{\mathcal{T}_r}
 %  %  R_{\left(h_i, t_i\right)}
 %    =\mathtt{Mean}(\{\mathtt{RML}(f({h}),f({t});\Phi)\mid \nonumber \\ (h,r,t)\in \mathcal{S}_r\}),
 %    \label{eqn:relation-meta}
 % \end{align}
where $f(\cdot)$ is a pre-trained encoder that generates $d$-dimensional embeddings for the entities. Moreover,
$\mathtt{RML}$ is the \emph{relation-meta learner}, implemented as a two-layer multi-layer perceptron (MLP).

It is worth noting that the prior $\Phi$ initializes the relation-meta learner $\texttt{RML}$, which further generates the relation meta $R_{\mathcal{T}_r}$. Subsequently, $R_{\mathcal{T}_r}$ is rapidly updated by a \emph{gradient meta} $G_{\mathcal{T}_r}$ calculated from the loss on the support set, i.e., $R'_{\mathcal{T}_r}=R_{\mathcal{T}_r}-\beta G_{\mathcal{T}_r}$, where $\beta$ is the step size. The resulting $R'_{\mathcal{T}_r}$ serves as the relation meta adapted to the support set, which is used to calculate the loss on the query set. Finally, the query loss is backpropagated to update the prior $\Phi$ and the embedding matrix $\texttt{emb}$ for the entities. %which contains solely the learnable embedding of entities. 

More specifically, the support and query losses are calculated as follows. 
\begin{align}
    L_{\mathcal{S}_r}=\textstyle\sum_{(h,r, t) \in \mathcal{S}_r}[\gamma+s(\mathtt{emb}(h), R_{\mathcal{T}_r}, \nonumber \\\mathtt{emb}({t}))
    -s(\mathtt{emb}(h),R_{\mathcal{T}_r},\mathtt{emb}(t'))]_{+}\label{eqn:supploss}\\
    L_{\mathcal{Q}_r}=\textstyle\sum_{(h,r, t) \in \mathcal{Q}_r}[\gamma+s(\mathtt{emb}(h),R'_{\mathcal{T}_r}, \nonumber \\\mathtt{emb}(t))
    -s(\mathtt{emb}(h),R'_{\mathcal{T}_r},\mathtt{emb}(t'))]_{+}\label{eqn:queryloss}
\end{align}
Here, $s(\cdot,\cdot)$ is a scoring function such as TransE \cite{bordes2013translating}, i.e., $s(\mathbf{h},\mathbf{r}, \mathbf{t})=\|\mathbf{h}+\mathbf{r}-\mathbf{t}_i\|$, where $\|\cdot\|$ denotes the $L_2$ norm. 
$\mathtt{emb}$ is an entity embedding matrix, which is initialized by a pre-trained model and can be further optimized during meta-training. 
% that further optimizes the pre-trained entity embeddings. 
$(h,r, t')$ refers to a negative triplet of the relation $r$, where $t'$ is randomly sampled from the type-constrained candidate set. $\gamma$ represents the margin which is a hyper-parameter. $[\cdot]_{+}$ denotes the positive part of the input value.

\stitle{Meta-testing.}
The meta-testing stage follows the same pipeline as meta-training, except that 
we cannot compute and backpropagate the query loss to update model parameters ($\Phi$ and $\texttt{emb}$). For each novel task $\mathcal{T}_r \in \mathcal{D}^\text{te}$, like meta-training, we first generate the relation meta $R_{\mathcal{T}_r}$ using the prior $\Phi$, then update it as $R'_{\mathcal{T}_r}$ based on the support set. Then, for each partial triplet $(h_j,r,?)$ in the query set, we rank the candidate entities $\mathcal{C}_{h_j,r}$ in ascending order of the scoring function $s(h_j, R'_{\mathcal{T}_r}, t_j)$ for every $t_j\in \mathcal{C}_{h_j,r}$.

% \stitle{Entity-Context}
\begin{figure*}[t]
    \hspace*{-10pt}
    \centering
    \includegraphics[width=1.1\linewidth]{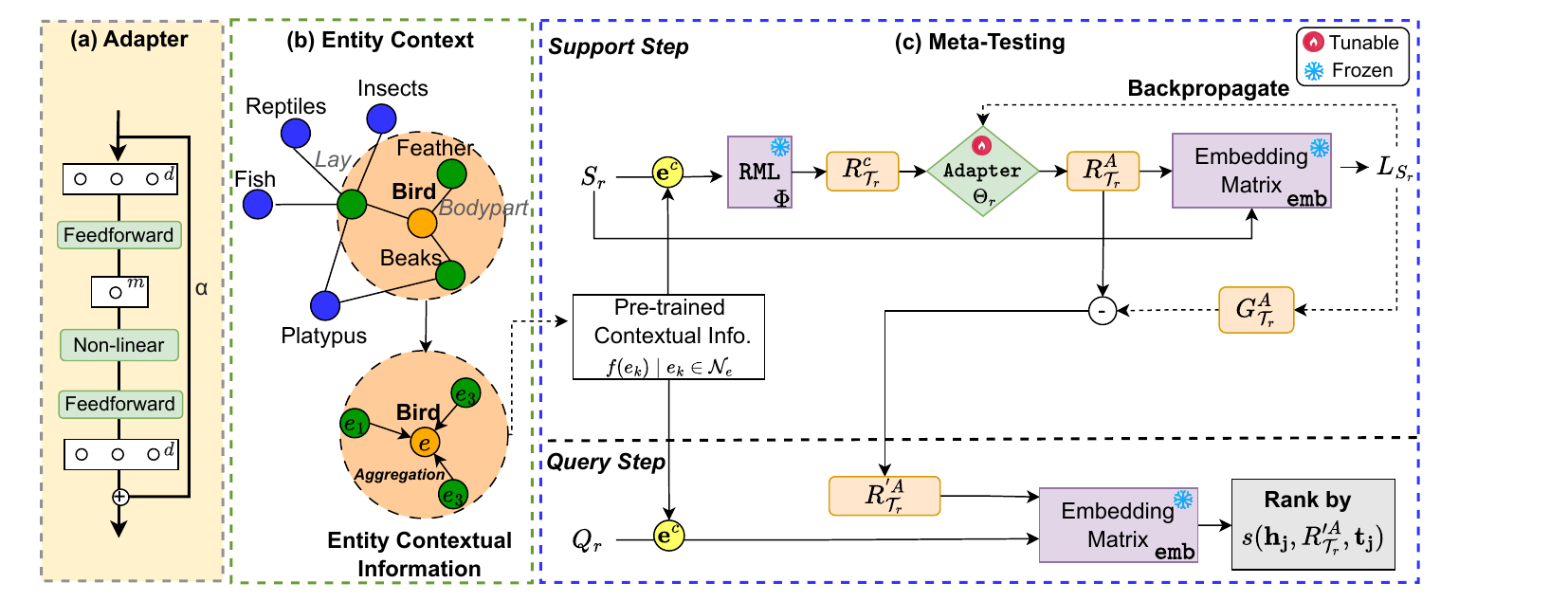}%
    % \vspace{-25mm}
    \caption{Illustration of key concepts in \model, hinging on an entity-aware adapter (a, b) in the meta-testing stage (c). Note that we omit the meta-training stage, which is similar to meta-testing but with backpropagation of the query loss to update the model parameters ($\mathtt{emb}$ and $\Phi$). }
    \label{fig:framework}
\end{figure*}

\section{Methodology} 

In this section, we introduce the proposed approach \model. As depicted in Fig.~\ref{fig:framework}, \model~has two important components, namely, (a) adapter network and (b) entity context. On one hand, the adapter network aims to facilitate relation-specific and tunable adaptation of meta-learned prior in a parameter-efficient manner. On the other hand, the entity context aims to enrich the adapter with contextual information pertinent to the target relation, enabling more precise adaptation to each distinct relation. 
The two components are integrated into a \emph{context-aware adapter}, which enhances FSRL for the novel relations in meta-testing.

In the rest of the section, we first introduce the context-aware adapter, followed by the details of the meta-training and meta-testing stages.

\subsection{Context-aware Adapter}\label{sec:method:adapter_architecture}  

We enhance the adaptation to novel relations at the model level through an adapter module, and at the data level through context-aware adaptation. 

% \textbf{\textit{Detailed information of \texttt{Adapter} sturcture.}} 
\stitle{Adapter.}
The objective of the adapter is to achieve a relation-specific adaptation for the novel relations in meta-testing, to overcome the divergence from the seen relations in meta-training. Specifically, we adapt the relation meta $R_{\mathcal{T}_r}$ to the target relation $r$ through the adapter module, as follows.
\begin{align}\label{eqn:adapter}
%\small{
    R^A_{\mathcal{T}_r} &= \texttt{Adapter}(R_{\mathcal{T}_r};\Theta_r)  \nonumber\\
    &= \alpha\cdot \texttt{FFN}(R_{\mathcal{T}_r};\Theta_r)  + (1-\alpha)\cdot R_{\mathcal{T}_r}
    %}
\end{align}
where the output from the adapter is $R^A_{\mathcal{T}_r}$, the \emph{adapted} relation meta specific to the relation $r$. 
The adapter module consists of a lightweight feed-forward network (\texttt{FFN}) and a residual layer, as shown in Fig.~\ref{fig:framework}(a), where $\alpha$ is a hyper-parameter to balance the \texttt{FFN} and the residual, and $\Theta_r$ is the $r$-specific parameter of the adapter. 
%\lr{The adapter takes in unrefined relation meta $R_{\mathcal{T}_r}$ and outputs the adapted relation meta $R^A_{\mathcal{T}_r}$ for each task relation $r$}. 
Note that the \texttt{FFN} typically adopts a bottleneck structure, which projects the input dimension $d$ into a smaller dimension $m$, reducing the number of parameters to achieve parameter-efficient adaptation.

\stitle{Context-aware adaptation.}
At the data level, we augment the embedding of each entity (head or tail) by additional pre-trained contextual information from their related entities, as shown in Fig.~\ref{fig:framework}(b). The contextual information enables more tailored adaptation to each distinct novel relation.
\begin{align}\label{eqn:contextadapter}
 \mathbf{e}^c =\ &\textstyle\mu\cdot\texttt{Mean}(\{f(e_k)\mid e_k\in\mathcal{N}_e\}) \nonumber \\
    &+ (1-\mu)\cdot\mathtt{emb}(e)
\end{align}
where $e\in\mathcal{V}$ is an entity, $N_{e}$ is the set of neighbors of $e$ in the knowledge graph (without the novel relations in meta-testing), and $\mu$ is a hyper-parameter. 
For each neighbor $e_k$ of $e$, we utilize the pre-trained encoder $f(\cdot)$ to extract its embedding. 
In summary, the input is the original entity embedding $\mathtt{emb}(e)$ and the mean contextual embedding aggregated from its neighbors $\mathcal{N}_e$, and the output is the augmented entity embedding $\mathbf{e}^c$. In this way, the augmented embedding $\mathbf{e}^c$ preserves the embedding trained via $\texttt{emb}$, while leveraging pre-trained graph contextual information.

Given the context-augmented entity embeddings $\mathbf{e}^c$ , we can derive the \emph{context-aware} relation meta, $R^c_{\mathcal{T}_r}$, by rewriting Eq.~\eqref{eqn:relation-meta} as follows.
\begin{align}
    R^c_{\mathcal{T}_r}
  %  R_{\left(h_i, t_i\right)}
    =\mathtt{Mean}(\{\mathtt{RML}(\mathbf{h}^c,\mathbf{t}^c;\Phi)\mid \mathcal{S}_r\})
    \label{eqn:relation-meta-context}
 \end{align}
 % \begin{align}
 %    R^c_{\mathcal{T}_r}
 %  %  R_{\left(h_i, t_i\right)}
 %    =\mathtt{Mean}(\{\mathtt{RML}(\mathbf{h}^c,\mathbf{t}^c;\Phi)\mid (h,r,t)\in \mathcal{S}_r\}).
 %    \label{eqn:relation-meta-context}
 % \end{align}
Subsequently, we update the adapter in Eq.~\eqref{eqn:adapter} to take in the context-aware relation meta, such that $R^A_{\mathcal{T}_r} = \mathtt{Adapter}(R^c_{\mathcal{T}_r};\Theta_r)$.

% the context-aware relation meta $R^c_{\mathcal{T}_r}$ passes through the adapter module $R^A_{\mathcal{T}_r} = \mathtt{Adapter}(R^c_{\mathcal{T}_r};\Theta_r)$ to output the relation-specific and context-aware relation meta $R^A_{\mathcal{T}_r}$ to be further utilized in meta-testing.}

\subsection{Integration with Meta-learning}\label{sec:method:meta-learning}  

Following MetaR \cite{chen2019meta}, our approach \model\ consists of the meta-training and meta-testing stages. 
%\lr{During meta-testing, we explain the implementation of context-aware adapter and introduce how it is being learned.}
%\zz{We will also introduce xxx(Adaptation, Adater tuning...) in this subsection.}

\stitle{Meta-training.} Our approach focuses on relation-specific adaption on novel relations in meta-testing, closing their gap from the seen relations due to divergence in distributions. Hence, our meta-training stage largely follows that of MetaR, as described in Sect.~\ref{sec:prelim}. The only difference is that, to have a consistent architecture to our meta-testing stage, we also include an adapter module in meta-training. Due to space constraint, we detail the description of the meta-training process in Appendix~\ref{meta-train}.

\stitle{Meta-testing.}  
% Following MetaR, the model parameters (i.e. entity embedding $e_{i}$ and relation meta-learner) are imported from the meta-train stage earlier. 
The process is outlined in Fig.~\ref{fig:framework}(c).
The meta-training stage learns the  embedding matrix $\texttt{emb}$ and prior $\Phi$ from the tasks on seen relations. To transfer the prior $\Phi$ to novel relations in meta-testing, it undergoes adaptation from two perspectives. As in traditional meta-learning, one form of  adaptation involves a rapid fine-tuning step on the support set. However, meta-learning assumes that the seen and unseen relations are drawn from an identical distribution, but the distribution of a novel relation might diverge significantly from those of the seen relations. Hence, we propose an additional form of adaptation based on the context-aware adapter in Sect.~\ref{sec:method:adapter_architecture}.

In the following, we elaborate on the adaption process and how the adapter is tuned, and outline the overall algorithm for meta-testing.

\stitle{Adaptation.}
On one hand, our adapter module is used to transform the context-aware relation meta $R^{c}_{\mathcal{T}_{\text{r}}}$, which is generated from the global prior $\Phi$ via Eq.~\eqref{eqn:relation-meta-context}, into a locally adapted version, i.e., $R^{A}_{\mathcal{T}_{\text{r}}}= \mathtt{Adapter}(R^{c}_{\mathcal{T}_{\text{r}}};\Theta_r)$. On the other hand, the local relation meta is further adapted by a quick gradient step on the support set, following conventional MAML-based adaptation.
\begin{align}
    G^{A}_{\mathcal{T}_r}&=\nabla_{R^{A}_{\mathcal{T}_r}} L_{\mathcal{S}_{r}},\label{eq:gradient-context} \\
    R'^{A}_{\mathcal{T}_r} &= R^{A}_{\mathcal{T}_r}-\beta G^{A}_{\mathcal{T}_r}. \label{eq:relation-meta-final}
\end{align}

\stitle{Adapter tuning.}
In the meta-testing stage, the global prior $\Phi$ and embedding $\texttt{emb}$ learned from meta-training are frozen,
while only the adapter, parameterized by $\Theta_{r}$, is optimized in a task-wise manner for $r$-specific adaptation. 
First, $\Theta_{r}$ is randomly initialized for each unseen task $\mathcal{T}_r\in \mathcal{D}^\text{te}$ without leveraging any meta-learned knowledge, to overcome the potential divergence from seen tasks in  $\mathcal{T}_r\in \mathcal{D}^\text{tr}$.
Next, $\Theta_{r}$ is optimized on the support set $\mathcal{S}_r$, using the same support loss in Eq.~\eqref{eqn:supploss}. In other words, gradients on the support loss is backpropagated to update $\Theta_r$ within each task $\mathcal{T}_r$.

\stitle{Scoring.} The final adapted relation meta from Eq.~\eqref{eq:relation-meta-final}, $R'^{A}_{\mathcal{T}_r}$, is used on the query set, to score and rank the candidates for the missing tail entities. The scoring function follows that of MetaR in Sect.~\ref{sec:prelim}. That is, for a given head $h$ from the query set $\mathcal{Q}_r$, for each candidate tail $t\in\mathcal{C}_{h,r}$, we compute 
$s(\mathtt{emb}(h),R'^{A}_{\mathcal{T}_r},\mathtt{emb}(t))=\|\mathtt{emb}(h)+R'^{A}_{\mathcal{T}_{\text{r}}}-\mathtt{emb}(t))\|$, and rank the candidates in $\mathcal{C}_{h,r}$ by the computed score.

\stitle{Algorithm.}
We outline the algorithm of our meta-testing procedure in Algorithm~\ref{algo}. 
Compared to MetaR, the novel components in \model~ constitute the injection of contextual information specific to each target relation (lines 2--3), and
the insertion of the adapter module (lines 5--8), which are integrated into a context-aware adapter. Despite the additional components, the overall time complexity remains unchanged and is linear to the number of shots. The overhead of the adapter module is negligible due to its parameter-efficient design.

\begin{algorithm}[hbt!]
\small
	\caption{Meta-testing for \model} \label{algo}
	\begin{algorithmic}[1]
    \Require Few-shot tasks for novel relations $\mathcal{D}^{\text{te}}$, embedding matrix $\texttt{emb}$, prior $\Phi$
    % , relation-specific $\Theta_\texttt{r}$
        
    % model $\Theta_{\text{tr}}=\{\Theta_\texttt{E}, \Theta_\texttt{r}\}$ and randomly initialized adapter parameters $\Theta^{r}_{\texttt{A}}$
		\For {each task $\mathcal{T}_r=(S_r, Q_r) \in \mathcal{D}^{\text{te}}$}
        \State Compute context-augmented entity embeddings, \hspace*{3.5mm} $\{\mathbf{e}^c\}$, from Eq.~\eqref{eqn:contextadapter};
		\State Compute context-aware relation meta, $R^{c}_{\mathcal{T}_r}$, from \hspace*{3.8mm} Eq.~\eqref{eqn:relation-meta-context};
        \While {$\Theta_{r}$ not converged}
        \State Compute adapted relation meta, $R^{A}_{\mathcal{T}_r}\gets\hspace*{8.5mm} \mathtt{Adapter}(R^c_{\mathcal{T}_r};\Theta_r)$;
        \State Compute support loss $L_{S_r}$ by Eq.~\eqref{eqn:supploss};
        \State Compute $G^{A}_{\mathcal{T}_r}$, the gradient of  $R^{A}_{\mathcal{T}_r}$ by Eq.~\eqref{eq:gradient-context};
        \State Update adapted relation meta, $R'^{A}_{\mathcal{T}_r} \gets R^{A}_{\mathcal{T}_r}-\hspace*{8.5mm}\beta G^{A}_{\mathcal{T}_r}$ by Eq.~\eqref{eq:relation-meta-final};      
        \State Update $\Theta_{r}$ w.r.t.~$L_{S_r}$;
        \EndWhile

        \For {each $(h,r,?) \in \mathcal{Q}_r$}
        \State Rank candidates in $\mathcal{C}_{h,r}$ by the scoring function;
        \EndFor
        \EndFor
	\end{algorithmic} 
\end{algorithm} 

% Due to space constraint, we detail the description of the meta-training process in \textcolor{purple}{Appendix D}.

\section{Experiments}\label{sec:expt}
In this section, we conduct comprehensive experiments on our proposed approach \model.
% \footnote{Anonymized codes are included in Supplementary Material for review.}.
%, including performance comparison, 
% we evaluate the performance of our proposed method through comparisons against the state-of-the-art baselines on FSRL datasets, 
% followed by detailed 
%ablation study, efficiency analysis, and sensitivity analysis.

\subsection{Experiment Setup}
\stitle{Datasets.} We utilize three benchmark datasets, namely, WIKI, FB15K-237 and UMLS. Table~\ref{table:1}  depicts the dataset details, including the pre-train/train/validation/test splits on the relations. The pre-trained encoder, $f(\cdot)$, which provides initial entity embeddings for the FSRL models, are based on the pre-train split.
Note that the four splits are mutually exclusive to avoid information leakage \cite{zhang2020few}. 
%the relations in test split are unseen from pre-train, train and validation. 
Additional details for the dataset can be found in Appendix~\ref{dataset}.

\begin{table}[t]
%\vspace{-7mm}
\caption{Statistics of datasets.}
\vspace{-2mm}
\centering
\label{table:1}
\addtolength\tabcolsep{-.5mm}
% \scriptsize
\resizebox{1\linewidth}{!}{%
\begin{tabular}{@{}c|r|r|rrrrr@{}}
\toprule
%& & & \multicolumn{4}{c}{\# Positive triplets} \\ % center-set header entries
& \multirow{2}{*}{Entities} & \multirow{2}{*}{Triplets} & \multicolumn{5}{c}{Relations} \\
&  &  & \small Total & \small Pre-train & \small Train & \small Valid & \small Test \\
\midrule
WIKI     & 4,838,244 &  5,859,240 & 639 & 456 & 133   & 16 & 34 \\
FB15K-237     & 14,541 &  281,624 &  237 & 118 &  75 & 11 & 33\\
UMLS    & 135 &  6529 &  25 & 5 &   10  & 5 & 5\\
\bottomrule
\end{tabular}}
\end{table}

\stitle{Metrics.} We employ two popular evaluation metrics, mean reciprocal rank (MRR) and hit ratio at top $N$ (Hit@$N$) to compare our approach against the baselines. Specifically, MRR reflects the absolute ranking of the first relevant item in the list and Hits@$N$ calculates the fraction of candidate lists in which the ground-truth entity falls within the first $N$ positions. %We set $N$ to 10.

\begin{table*}[t]
\centering
\caption{Performance comparison against baselines in the 3-shot setting. (Best: bolded, runners-up: underlined). %The improvements are calculated for our model \model\ relative to the next best model.
}
\vspace{-2mm}
\label{table:2} 
\small
\addtolength{\tabcolsep}{-1mm}
%\resizebox{1.0\textwidth}{!}{%
\hspace{-2mm}
 \fontsize{9.2}{9.2}\selectfont
\begin{tabular}{@{}c|ccc|ccc|ccc@{}} 
  \toprule
  &\multicolumn{3}{c|}{WIKI} & \multicolumn{3}{c|}{FB15K-237} & \multicolumn{3}{c}{UMLS} \\
  \cmidrule(lr){2-4} \cmidrule(lr){5-7} \cmidrule(lr){8-10}
  Models & MRR   & Hit@10 & Hit@1  & MRR  & Hit@10 & Hit@1  & MRR  &  Hit@10 & Hit@1\\
  \midrule
  TransE  & .031$\pm$.007  & .043$\pm$.012 & .021$\pm$.014 & .294$\pm$.005 & .437$\pm$.011 & .204$\pm$.014 & .178$\pm$.036   &   .310$\pm$.051 &  .146$\pm$.068\\
  DistMult   & .047$\pm$.003  & .082$\pm$.009 & .031$\pm$.011 &   .234$\pm$.008  & .364$\pm$.007& .208$\pm$.010 & .231$\pm$.035   & .337$\pm$.049 & .214$\pm$.067  \\
  ComplEx  & .093$\pm$.004 & .166$\pm$.011 & .071$\pm$.012 &  .239$\pm$.007  & .359$\pm$.010 & .205$\pm$.013 & .251$\pm$.038  & .351$\pm$.041 & .227$\pm$.058 \\
   RGCN  & .217$\pm$.012 & .363$\pm$.023 & .188$\pm$.031 & .332$\pm$.011 &   .495$\pm$.013  &  .241$\pm$.031 & .409$\pm$.059  & .549$\pm$.072 & .389$\pm$.089 \\
\midrule

% {\scriptsize{Few-shot}}
  GMatching  & .133$\pm$.017 & .331$\pm$.013 &.114$\pm$.026 & .309$\pm$.019     & .441$\pm$.015 & .245$\pm$.019 & .296$\pm$.059 & .532$\pm$.040 & .257$\pm$.087  \\
   FSKGC  & .131$\pm$.003  & .267$\pm$.010 & .104$\pm$.016  & .355$\pm$.005   & .523$\pm$.004  & .217$\pm$.011  &  .525$\pm$.031 & .682$\pm$.024 & .490$\pm$.038 \\ 
   GANA   & .291$\pm$.014  & .384$\pm$.012 & .272$\pm$.015 & \underline{.388}$\pm$.004  & .553$\pm$.008 & \textbf{.301}$\pm$.017  & .541$\pm$.045  & .721$\pm$.076 & .502$\pm$.047\\
   FAAN   & .278$\pm$.018 & .421$\pm$.020 & .275$\pm$.024 & .363$\pm$.009  & .542$\pm$.007 & .279$\pm$.013 & .545$\pm$.034   & .746$\pm$.120 & .505$\pm$.068\\
   HiRe  & .300$\pm$.028  & \underline{.444}$\pm$.012 & \underline{.282}$\pm$.015& .378$\pm$.013 & \underline{.571}$\pm$.011 & .281$\pm$.015 & \underline{.577}$\pm$.060  & \underline{.752}$\pm$.066 & \underline{.533}$\pm$.089\\
 MetaR   & \underline{.314}$\pm$.013  & .420$\pm$.016 & .274$\pm$.028 & .368$\pm$.007  & .536$\pm$.005 & .251$\pm$.012 & .435$\pm$.075   & .601$\pm$.095 & .417$\pm$.103\\
\midrule
 \model & \textbf{.347}$\pm$.006 &  \textbf{.454}$\pm$.012 & \textbf{.317}$\pm$.013  & \textbf{.405}$\pm$.012  & \textbf{.575}$\pm$.014 & \underline{.297}$\pm$.019 & \textbf{.608}$\pm$.067  & \textbf{.780}$\pm$.044 & \textbf{.555}$\pm$.062\\
\bottomrule
\end{tabular}%
\hspace{15mm}
% %}
\end{table*}

\stitle{Baselines.} \model~is compared with a series of baselines in two major categories.
(1) \emph{Supervised relation learning}. They learn one model for all the relations in a supervised manner. We choose four classic and popular supervised methods: \textbf{TransE} \cite{bordes2013translating}, \textbf{DistMult} \cite{yang2014embedding}, \textbf{ComplEx} \cite{trouillon2016ComplEx} and \textbf{RGCN} \cite{schlichtkrull2018modeling}. We follow the same setup in GMatching \cite{xiong2018one}, which trains on the triplets combined from the pre-train and train splits, as well as the support sets of the test splits. 
% which operates on heterogeneous graphs with multiple relation types
%. Such traditional supervised models often require adequate training triplets for each relation task to learn static representations of entities and relations. 
(2) \emph{Few-shot relation learning (FSRL).} They are designed for few-shot relation prediction tasks. We choose several state-of-the-art FSRL methods, as follows: \textbf{GMatching} \cite{xiong2018one}, \textbf{FSKGC} \cite{zhang2020few}, \textbf{GANA} \cite{niu2021relational}, \textbf{FAAN} \cite{sheng2020adaptive}, \textbf{HiRe} \cite{wu2023hierarchical}, \textbf{MetaR} \cite{chen2019meta} and the details can be found in Appendix~\ref{baselines}.

% \textbf{GMatching} \cite{xiong2018one} uses a neighbor encoder and a matching network, assuming that all neighbors contribute equally. \textbf{FSKGC} \cite{zhang2020few} encodes neighbors with a fixed attention mechanism, and applies a recurrent autoencoder to aggregate the few-shot instances in the support set. \textbf{GANA} \cite{niu2021relational} improves on FSKGC by having a gated and attentive neighbor aggregator to capture valuable contextual semantics of each few-shot relation. \textbf{FAAN} \cite{sheng2020adaptive} introduces an adaptive neighbor encoder for different relation tasks. \textbf{HiRe} \cite{wu2023hierarchical} brings in a hierarchical relational learning framework which considers triplet-level contextual information in contrastive learning. \textbf{MetaR} \cite{chen2019meta} utilizes a MAML-based framework, which aims to learn a good initialization for the unseen relations, followed by an optimization-based adaptation. 

\stitle{Implementation Details.}
For a fair comparison, we initialize all FSRL models with pre-trained entity and relation embeddings, if needed. Other details can be found in Appendix~\ref{Implementation_details}.

\subsection{Comparison with Baselines}
Table~\ref{table:2} reports the quantitative comparison of \model\ against other baselines in the 3-shot setting, i.e., the size of the support set is 3. (We study the effect of the number of shots in Sect.~\ref{sec:expt:sensitivity}). 

Overall, our model \model\ outperforms other baselines across all the three datasets, which demonstrates the benefit of incorporating context-aware adapter for FSRL. In particular, \model\ outperforms the most competitive baseline HiRe by 9.84\% in terms of average MRR and 2.22\% in terms of average Hit@10.
Furthermore, we also draw the following observations. 

First, \emph{supervised relation learning} methods tend to perform worse as compared to \emph{FSRL} methods as they are not designed to handle novel relations in few-shot setting. Meanwhile, as RGCN considers the neighborhood aggregated information, it consistently outperforms other supervised relation learning models across all the three datasets. 

Second, among the \emph{FSRL} methods, GMatching is designed for one-shot setting, and a simple mean pooling is applied to handle multiple shots, resulting in unsatisfactory performance. FSKGC generally performs better than GMatching as it extends the one-shot setting to more shots and explores new ways to encode neighbors with an attention mechanism. Although both GMatching and FSKGC consider neighborhood information, the simple neighborhood aggregation design is not expressive enough to capture complex relations. On the other hand, GANA and FAAN outperform GMatching and FSKGC as they consider the neighborhood information via more expressive neighborhood aggregators, employing a gated attentive neighborhood aggregator and a task-specific entity adaptive neighbor encoder, respectively. On top of the neighborhood information, HiRe considers triplet-level contextual information to generalize to few-shot relations, thus outperforming GANA and FAAN in most cases. 

Third, MetaR is the closest to \model\ in terms of the model architecture. 
In particular, the addition of context-aware adapter in the meta-learning framework enables more precise and relation-specific adaptation to each novel relation, mitigating the distribution shift between different relations.  
Hence, compared against MetaR, \model\ achieves an average improvement of 20.1\% in MRR and 15.1\% in Hit@10.
%to deal with data distribution shift has led to the marginal improvement against HiRe, achieving a MRR gain of 15.7$\%$, 10.1$\%$ and 3.72$\%$ for WIKI, FB15K-237 and UMLS dataset. 

% While the architecture of \model\ cannot consider imaginary embedding in ComplEx~\cite{trouillon2016ComplEx} and RotatE~\cite{sun2019rotate} scoring functions, additional experiments have been conducted on other scoring functions such as DistMult~\cite{yang2014embedding}, RESCAL~\cite{nickel2011three} and HolE~\cite{nickel2016holographic} found in \textcolor{purple}{Appendix E.2})

\begin{table}[t]
\centering
\caption{Ablation study under the 3-shot setting. A: adapters in both meta-train and meta-test stage; C: using contextual information; A$^\text{tr}$: adapter in meta-training; A$^\text{te}$-Trf: adapter in meta-testing by transferring from meta-trained adapter. MetaR is considered a special variant without any of these components.}
\label{table:3} 
\small
\fontsize{9.2}{9.2}\selectfont
\begin{tabular}{c|ccc} 
  \toprule
   & \multicolumn{3}{c}{MRR} \\   \cmidrule(lr){2-4} 
   &  WIKI  &  FB15K-237  & UMLS\\  \midrule
    MetaR &  .314$\pm$.013 &  .368$\pm$.007 &   .435$\pm$.075  \\\midrule
 W/o A &   .312$\pm$.004 &    .385$\pm$.007 &     .389$\pm$.036    \\
    W/o C &  .332$\pm$.009 &   .395$\pm$.007 &   .590$\pm$.043     \\
W/o A$^\text{tr}$  &   .343$\pm$.015 &   .395$\pm$.012 &  .583$\pm$.047    \\
 A$^\text{te}$-Trf  & .341$\pm$.008  &  .394$\pm$.014 &    .586$\pm$.046   \\
 \midrule
   \model\   & \textbf{.347}$\pm$.006  &  \textbf{.405}$\pm$.012  & \textbf{.608}$\pm$.067  \\  
\bottomrule
\end{tabular}
\end{table}

\subsection{Ablation Study}\label{ablation}
To investigate the impact of various modules, we study four variants of our model, as shown in Table~\ref{table:3}.
(1) 
\textbf{W/o A:} We remove the adapter module entirely, while retaining the contextual information for entities. It shows a pronounced drop in performance and can be worse than MetaR, implying that simply adding contexts without further adaptation may introduce additional noises that harm the performance. In particular, the effect is the most significant on the UMLS dataset, which could be attributed to its smaller size. Specifically, the meta-learned model is more likely to overfit to the smaller meta-training data, and thus it becomes more important to have the adapter to deal with the distribution shift from meta-training. (2) \textbf{W/o C:} we remove the contextual information, while retaining the adapter. The lower performance shows that leveraging contexts can enhance model performance, if adapter is also present. 
%More specifically, we have experimented on two example entities, `age\_group' and `family\_group', from the UMLS dataset to further illustrate how entity contexts could improve model performance in Appendix~\ref{casestudy}.
(3) \textbf{W/o A$^\text{tr}$:} We remove the adapter only from meta-training (see Appendix~\ref{meta-train}). % (A$^\text{tr}$). 
The decrease in performance justifies the need for a consistent architecture in both meta-training and -testing. 
%Results show that the existence of adapter network in meta-training benefits model performance. This reinforces the purpose of adding the adapter network to meta-training (see Appendix~\ref{meta-train}) for a consistent model architecture in both meta-training and meta-testing stages. 
(4) \textbf{A$^\text{te}$-Trf:} we transfer the meta-trained adapter parameters to meta-testing, serving as the initialization for the adapter in meta-testing. In contrast, in \model, the adapter is randomly initialized in meta-testing. This variants suffer from a notable drop in performance, which is consistent with our earlier hypothesis on the divergence between relations. Specifically, due to the distribution shift in novel relations encountered in meta-testing, using prior adapter parameters from meta-training may be counterproductive. 

\begin{figure*}[hbt!]
    \hspace{-2mm}
    \hspace{35mm}
    \begin{minipage}[t]{0.4\textwidth}
        \centering\hfill
        \includegraphics[width=\textwidth]{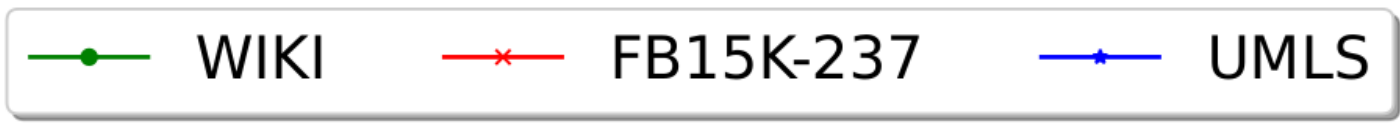}
        \hspace{-12mm}
    \end{minipage}
    
    \begin{minipage}[t]{.24\textwidth}
        \centering
        \includegraphics[width=\textwidth]{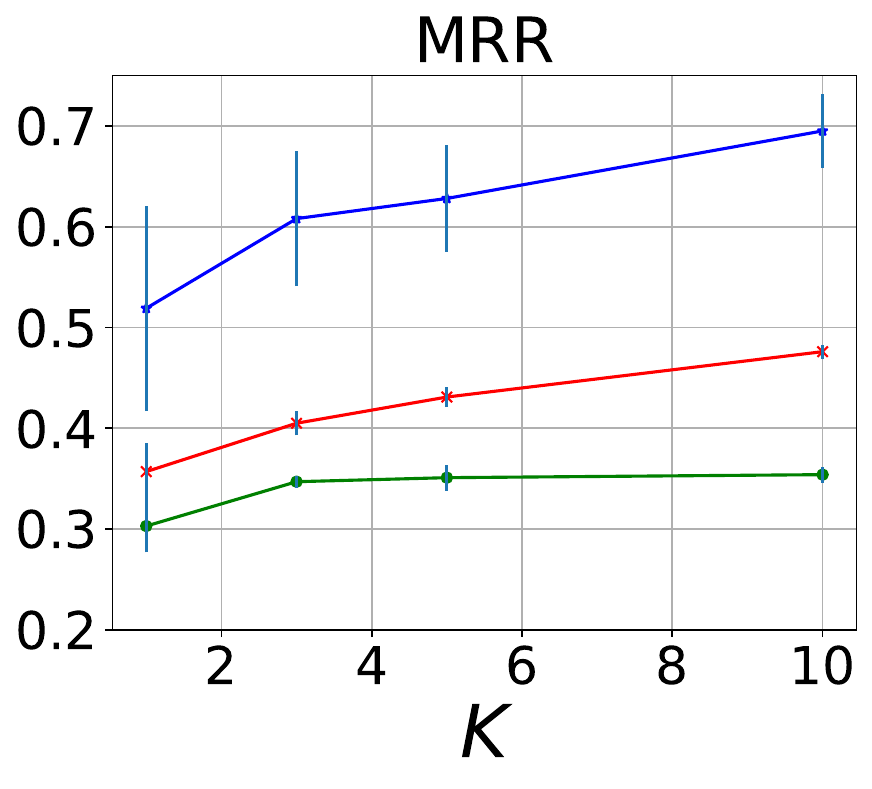}%
        \vspace{-1mm}
        \subcaption{Few-shot size}\label{fewshot}
    \end{minipage}
    \hspace{-1.5mm}
    \begin{minipage}[t]{.24\textwidth}
        \centering
        \includegraphics[width=\textwidth]{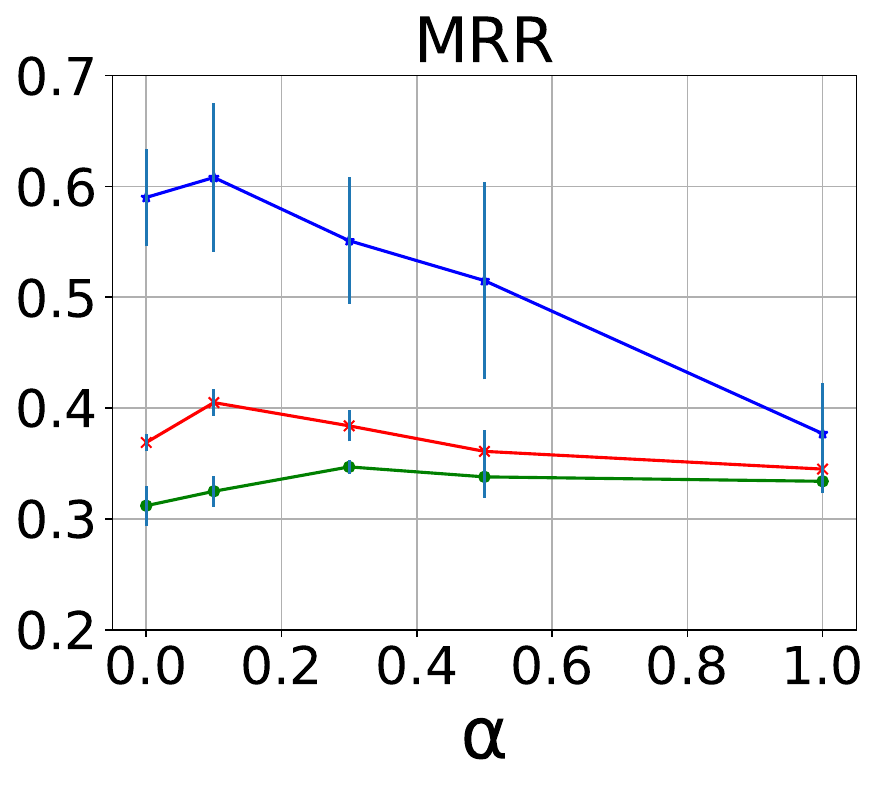}%
        \vspace{-1mm}
        \subcaption{Adapter ratio}\label{adaptercoeffcient}
    \end{minipage}  
    \begin{minipage}[t]{.24\textwidth}
        \centering
        \includegraphics[width=\textwidth]{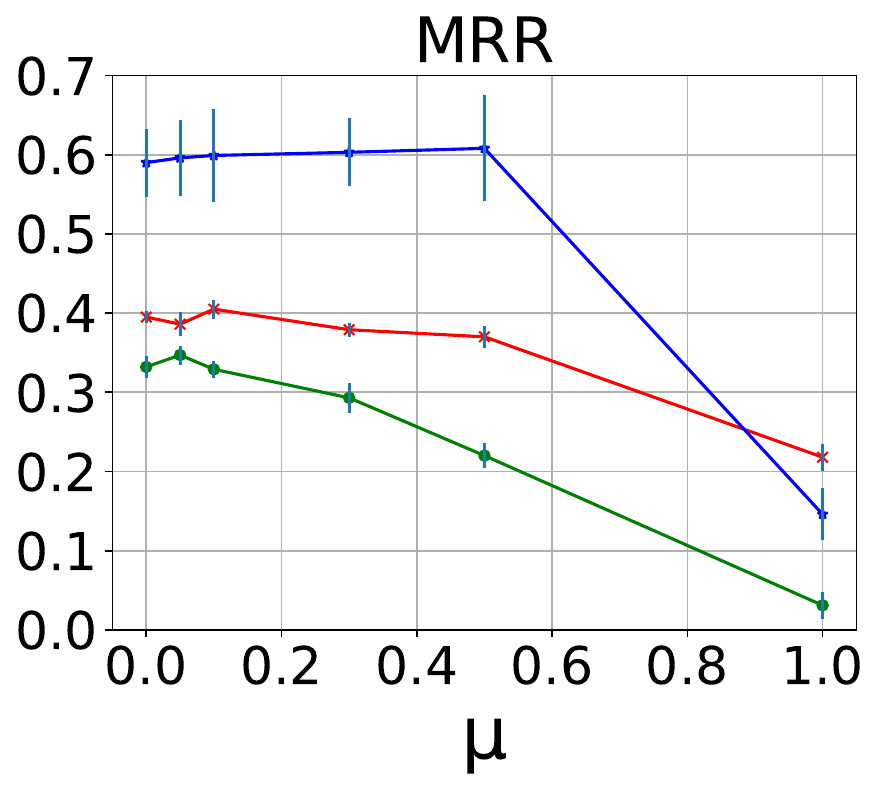}%
        \vspace{-1mm}
        \subcaption{Context ratio}\label{beta}
    \end{minipage}  
    \begin{minipage}[t]{.24\textwidth}
        \centering
        \includegraphics[width=\textwidth]{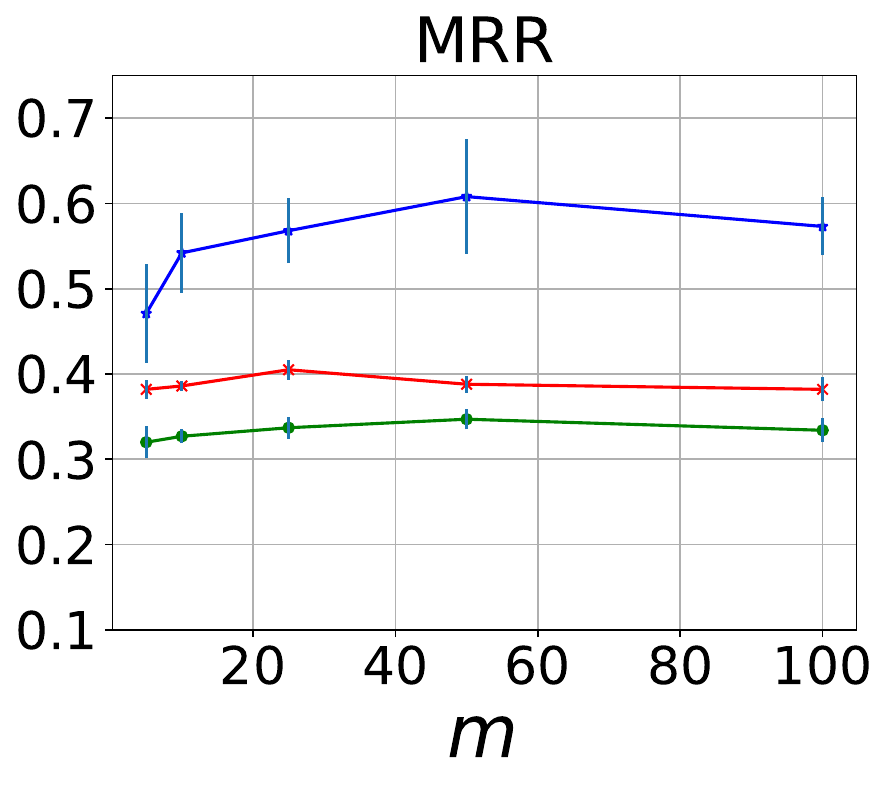}%
        \vspace{-1mm}
        \subcaption{Adapter neurons}\label{neurons}
    \end{minipage}
    \vspace{-2mm}
    \caption{Sensitivity analysis for the number of shots and hyperparameters.}
    \label{fig:1-2}
\end{figure*}

\subsection{Efficiency Analysis}
We analyze the parameter and runtime efficiency of our adapter module.
%, including its parameter efficiency and runtime costs. %followed by measuring the actual runtime for the meta-train and meta-test stage. Results demonstrate the parameter efficient design of the adapter module.

\begin{table}[t] 
\centering
\caption{Comparison of our adapter and MetaR in terms of number of parameters.}\label{table:parametersize}
\small
% \addtolength{\tabcolsep}{3.5mm}
\fontsize{9.2}{9.2}\selectfont
\begin{tabular}{c|ccc}
  \toprule
 % & \multicolumn{3}{c}{Number of parameters}  \\ \cmidrule{2-4}
 & WIKI & FB15K-237 & UMLS   \\  \midrule
   MetaR  &   241,967,556 & 1,650,206 & 234,306    \\
   Our adapter    &   5,125 & 5,125 & 5,125   \\
 \midrule
  \% of MetaR    &   0.002 & 0.311 & 2.187    \\
\bottomrule
\end{tabular}%
%}
\end{table}

\stitle{Parameter efficiency.} We first study the parameter overhead from the addition of adapter module. As shown in Table~\ref{table:parametersize}, the number of parameters in the adapter module is negligible w.r.t.~MetaR. The  parameter-efficient design implies that our adapter tuning is less likely to overfit to the few-shot examples. Also note that, compared to MetaR, the only new parameters of \model\ belong to the adapter module. 

\begin{table}[t]
\centering
\caption{Runtime (in seconds) for meta-training and meta-testing.}\label{table:runtime}
\addtolength{\tabcolsep}{-1mm}
% \small
\fontsize{9}{9}\selectfont
{\begin{tabular}{c|c|ccc}
  \toprule
Stage & Model & WIKI & FB15K-237 & UMLS   \\  \midrule
  & GMatching  &   28,941 & 19,678 & 12,061    \\
  & FSKGC  &   28,742 & 20,765 & 13,869    \\
  & GANA  &   35,374 & 28,167 & 15,081    \\
 Meta-train   & FAAN  &   32,036 & 23,675 & 11,302    \\
   (total) & HiRe  &   34,257 & 27,736 & 12,213    \\
  & MetaR  &   22,691 & 16,504 & 8,802    \\
  & \model    &   24,085 & 17,529 & 9,656   \\ \midrule
   & GMatching  &   0.009 & 0.004 & 0.039    \\
  & FSKGC  &   0.013 & 0.004 & 0.042    \\
  & GANA  &   0.017 & 0.006 & 0.064    \\
 Meta-test   & FAAN  &   0.016 & 0.005 & 0.053    \\
   (per instance) & HiRe  &   0.036 & 0.007 & 0.051    \\
 & MetaR  &   0.012 & 0.005 & 0.043    \\
 & \model    &   0.045 & 0.008 & 0.053   \\
\bottomrule
\end{tabular}}%
%}
\end{table}

\stitle{Runtime efficiency.} As reported in Table~\ref{table:runtime}, our approach \model\ generally incurs a lower or comparable total runtime for meta-training in comparison to various baselines. More specifically, compared to the most efficient approach MetaR, \model\ only takes slightly more time to complete the training. This observation shows that the parameter-efficient design of \model\ is able to achieve significant performance improvement with a decent runtime efficiency.

On the other hand, during meta-testing, we  observe a notable increase in the average runtime per instance due to the relation-specific adaptation in \model. However, it remains on a similar order of magnitude as the baselines, with only a marginal absolute difference, especially when considering the much longer meta-training stage.

% Next, we measure the runtime of meta-training and meta-testing stages. As shown in Table~\ref{table:runtime}, compared to MetaR, \model\ only incurs a marginal increase in the total runtime for meta-training. %,  taken after the insertion of the adapter module in $\model$. This observation shows that $\model$ is able to achieve performance improvement with minimal increase in runtime. 
% On the other hand, despite an increase in the average runtime per prediction during meta-testing, the absolute difference is minimal especially in comparison to the significantly longer runtime in the meta-training stage.

\subsection{Sensitivity Analysis}\label{sec:expt:sensitivity}
Lastly, we conduct a sensitivity analysis for various settings and hyperparameters in Fig.~\ref{fig:1-2}.
In particular, we vary the number of shots $K$ while fixing the hyperparamters, as well as several key hyperparameters while fixing $K=3$. 
In the figures, the $x$-axis refers to the range of $K$ or parameters against the MRR metric in $y$-axis. The error bars represent the spread of standard deviation for each data point.

\stitle{Few-shot size $K$.} The few-shot size $K$ refers to the number of triplets in the support set of each relation. As shown in Fig.~\ref{fig:1-2}(a), when $K$ increases, we consistently observe performance improvement as more data becomes available for training. %Therefore, we consistently observe performance improvement for MRR and an overall decrease in the corresponding standard deviation.

% \begin{figure}[hbt!]
%     \centering
%     \hspace{-33mm}
%     \includegraphics[width=7cm]{Figures/Lambda_MRR.pdf} 
%     \qquad
%     \hspace{-20mm}
%     \includegraphics[width=7cm]{Figures/Lambda_Hit10.pdf} 
%     \hspace{-33mm}
%     \caption{Impact of adapter Intensity, $\lambda$ }
%     \label{figure3}
% \end{figure}

\stitle{Adapter ratio $\alpha$.}
As given in Eq.~\eqref{eqn:adapter}, Sect.~\ref{sec:method:adapter_architecture}, \model\ employs a hyperparameter $\alpha$ which controls the weight of the residual in the context-aware adapter. A bigger $\alpha$ means less residual, giving the adapter more transformative power to adapt to each relation.  As shown in Fig.~\ref{fig:1-2}(b), the performance peaks when $\alpha$ = 0.1 on  FB15K-237 and UMLS, and $\alpha$ = 0.3 on WIKI. In general, [0.1, 0.3] appears as a robust range. Beyond the range, the performance starts to decrease, indicating that excessive adapter transformations are not beneficial. In particular, UMLS is more sensitive to the changes in $\alpha$. A potential reason is that UMLS is a relatively small dataset in a focused domain with less distribution shifts (see Fig.~\ref{fig:pilot}). Hence, it requires less adaptation across relations. %, and a stronger adapter may be counterproductive.
% a stronger adapter tends to cause overfitting to the support set, leading to unfavourable performance. % in query test set.

% \begin{figure}[hbt!]
%     \centering
%     \hspace{-33mm}
%     \includegraphics[width=7cm]{Figures/Beta_MRR.pdf} 
%     \qquad
%     \hspace{-20mm}
%     \includegraphics[width=7cm]{Figures/Beta_Hit10.pdf} 
%     \hspace{-33mm}
%     \caption{Impact of Contextual Coefficient, $\beta$}
%     \label{figure4}
% \end{figure}

\stitle{Context ratio $\mu$}. Similar to $\alpha$, $\model$ controls the weight of contextual information via the hyper-parameter $\mu$ in Eq.~\ref{eqn:contextadapter}, Sect.~\ref{sec:method:adapter_architecture}. As illustrated in Fig.~\ref{fig:1-2}(c), a smaller contextual ratio generally helps to improve the model performance across all datasets, while an excessively large ratio could bring in more noises and hurt the performance. In general, [0.1,0.3] appears to be a good range.  

% \begin{figure}[hbt!]
%     \centering
%     \hspace{-33mm}
%     \includegraphics[width=7cm]{Figures/Neuron_MRR.pdf} 
%     \qquad
%     \hspace{-20mm}
%     \includegraphics[width=7cm]{Figures/Neuron_Hit10.pdf} 
%     \hspace{-33mm}
%     \caption{Impact of No. of Neurons,  $M_s$ }
%     \label{figure6}
% \end{figure}

\stitle{Adapter neurons $m$.} We analyze how the number of neurons $m$ in the hidden layer of the adapter network affects model performance. Results in Fig.~\ref{fig:1-2}(d) show that the optimal $m$ is around 25 for FB15K-237, and 50 for WIKI and UMLS. Increasing $m$ further leads to a plateau in performance, suggesting the effectiveness of the bottleneck structure and ensuring a parameter-efficient design.

\stitle{Number of hops for contexts.} We investigate the impact of the number of hops considered in  neighborhood contexts, $\mathcal{N}_e$, in Eq.~\eqref{eqn:contextadapter}. Due to space limit, we present the results in Appendix~\ref{hops}.

\section{Conclusion}
In this paper, we proposed \model, a context-aware adapter for few-shot relation learning (FSRL). We investigated the limitation of  FSRL methods in prevailing meta-learning frameworks, which rely on the i.i.d.~assumption. This assumption may not hold for novel relations with distribution shifts from the seen relations. Based on this insight, we introduced a context-aware adapter module, enabling relation-specific, tunable and fine-grained adaptation for each distinct relation. Extensive experiments were conducted on three benchmark datasets, demonstrating the superior performance of \model.
%\ significantly outperforms state-of-the-art methods, but also offering an in-depth analysis of the strength and motivation of our method. 

% \lr{Despite the improvement in performances over state-of-the-art FSRL methods, \model\ is not designed to handle multi-modality knowledge graphs (MMKGs) \cite{liu2019mmkg}, which can be prevalent in the real-world. In future, we intend to explore how to effectively incorporate the use of MMKGs as contexts in our approach. Add more limitations for RelAdapter , not future work.}

%\clearpage
\section*{Limitations}
In \model, one potential limitation is that the context-aware adapter module is currently only integrated into the MetaR framework. Despite the significant improvement in performance, it would be ideal to explore the integration of \model\ with other meta-learning frameworks in general. 

\section*{Acknowledgement}
This research was supported by the Singapore Ministry of Education (MOE) Academic Research Fund (AcRF) Tier 1 grant (22-SIS-SMU-054) and National Research Foundation, Singapore under its AI Singapore Programme (AISG2-RP-2021-027).
Any opinions, findings and conclusions or recommendations expressed in this material are those of the author(s) and do not reflect the views of the Ministry of Education or National Research Foundation, Singapore. 
Ran Liu is also supported by the A*Star Graduate Scholarship offered by the Agency for Science, Technology and Research, Singapore for his PhD study.

% \lr{As \model\ is built on an existing meta-learning framework, MetaR \cite{chen2019meta}, the performance improvements against other baselines such as GMatching \cite{xiong2018one}, FSKGC \cite{zhang2020few} which do not share the same meta-learning architecture may not provide a fair comparison. Nevertheless, the significant performance improvement of \model\ when compared against MetaR does show the importance of introducing context-aware adapter module into the meta-learning framework. In the future, we may explore the implementation of \model\ on various backbones to investigate the impact on other model architectures.  }

\bibliography{Reladapter}

\appendix
\section*{Appendices}

\section{Meta-training stage}\label{meta-train}
\begin{figure*}[hbt!]
    \hspace*{-12pt}
    \centering
    \includegraphics[width=0.75\linewidth]{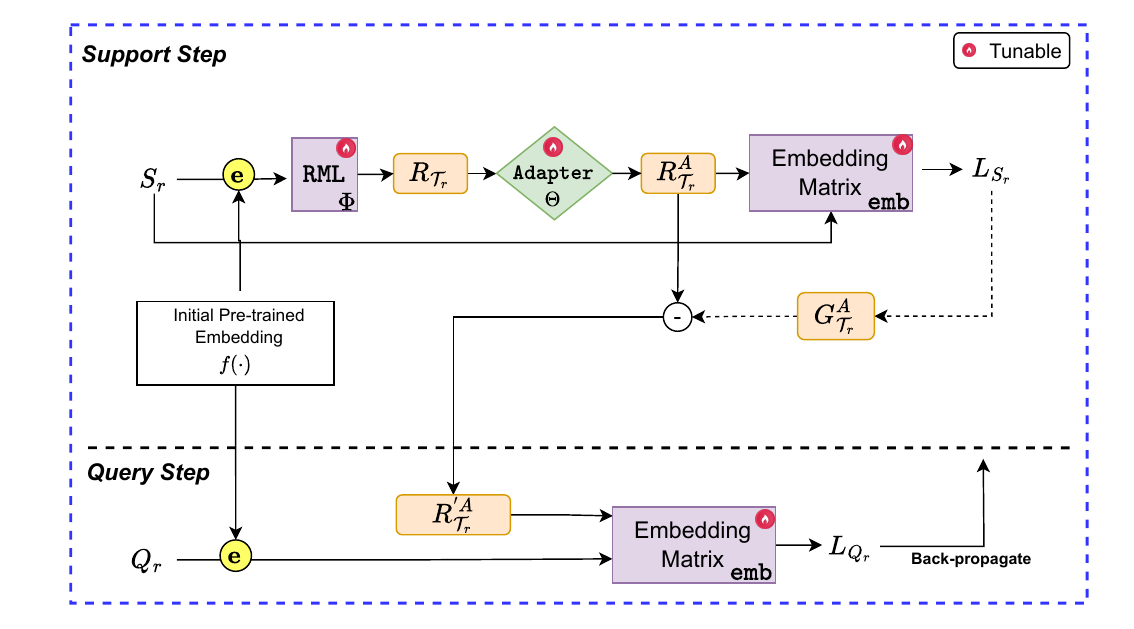}%
     \vspace{-3mm}
    \caption{Illustration of the meta-training stage.}
    \label{Meta-train_Framework}
\end{figure*}

In the meta-train stage as illustrated in Fig.~\ref{Meta-train_Framework}, we update the model with the query loss $\mathcal{L}_{\text{Qr}}$ in the same way as MetaR. In summary, we formulate the trainable parameters of the meta-train stage as $\Phi$, $\Theta$, $\texttt{emb}$
, where $\Phi$ is the set of trainable parameters in relation-meta learner \texttt{RML}, $\Theta$ is the set of trainable parameters of the adapter module and $\texttt{emb}$ is the embedding learner. The output of the meta-training stage includes the set of meta-trained entity embedding $\texttt{emb}$ as well as the relation-meta learner \texttt{RML}, which will be further used to adapt to downstream tasks in the meta-testing stage. The purpose of introducing the adapter network into the meta-training stage is to maintain consistency with the architecture of the meta-testing stage, which can improve performance. %More statistics can be found in ablation study of Section 5.3 from the main paper.

\section{Datasets}\label{dataset}
We utilize three benchmark datasets, namely, WIKI, FB15K-237 and UMLS. On each dataset, the relations are divided into four subsets: pre-training, training, validation and test, as shown in Table~\ref{table:1}. For the smaller dataset UMLS, all relations with less than 50 triplets are removed. % and the bottom 5 relations with the least number of triplets are used as few-shot relations. 

For supervised models not designed for FSRL (TransE, DistMult and ComplEx from OpenKE\footnote{https://github.com/thunlp/OpenKE}, and RGCN\footnote{https://github.com/JinheonBaek/RGCN}), we follow the same settings in GMatching \cite{xiong2018one} by using all triplets in the pre-train and train splits, as well as the support sets from the valid/test splits to train the models. For other FSRL models (GMatching\footnote{https://github.com/xwhan/One-shot-Relational-Learning}, FSKGC\footnote{https://github.com/chuxuzhang/AAAI2020$\_$FSRL}, MetaR\footnote{https://github.com/AnselCmy/MetaR}, GANA\footnote{https://github.com/ngl567/GANA-FewShotKGC}, FAAN\footnote{https://github.com/JiaweiSheng/FAAN}, HiRE\footnote{https://github.com/alexhw15/HiRe}), we follow the same FSRL splits as \model. All results are averaged among 5 runs.

Note that, as FB15K contain many inverse triplets which can cause leakage during training, it has been omitted and replaced with its subset FB15K-237 \cite{toutanova2015observed} which has all the inverse triplets removed. In addition, the popular WN18 and WN18RR datasets have also been omitted, as they contain insufficient relations and therefore are not suitable to be used in our experiments.

\label{Implementation_details}

\begin{table*}[t!]
\small
\caption{Tuned hyperparameter settings based on validation data.}
\vspace{-2mm}
\label{hyperparameters}
\begin{center}
\begin{tabular}{c| c c c c c} 
 \toprule
  & Hyperparameters & Range of values & Wiki & FB15K-237 & UMLS \\
 \midrule\midrule
 TransE & norm & 1,2 & 1 & 2 & 1 \\ 
 \midrule
DistMult & norm & 1,2  & 1 & 1 & 1 \\
\midrule
 ComplEx & norm & 1,2 & 1 & 1 & 1 \\
\midrule
 RGCN & dropout & 0.1,0.2,0.5 & 0.2 & 0.2 & 0.1 \\
\midrule
 Gmatching & aggregate & mean, max, sum & max & mean & max \\
\midrule
 MetaR & beta & 1,3,5,10  & 5 & 3 & 5 \\
\midrule
 FSRL & aggregate & mean, max, sum & max & max & max \\
\midrule
 GANA & beta & 1,3,5,10 & 5 & 5 & 5 \\
\midrule
 FAAN & dropout\_input & 0.1,0.3,0.5,0.8 & 0.5 & 0.3 & 0.5 \\
\midrule
 HIRE  & beta & 1,3,5,10 & 5 & 5 & 10 \\ 
\midrule
 \model\ & $\alpha$, $\mu$ & $0.05 \sim 1.0$ & $\alpha$ = 0.5, $\mu$ = 0.05 & $\alpha$ = 0.1, $\mu$ = 0.1 & $\alpha$ = 0.1, $\mu$ = 0.3 \\ 
 \bottomrule
\end{tabular}
\end{center}
\end{table*}

\section{Baselines}\label{baselines}
Few-shot relation learning (FSRL) are designed for few-shot relation prediction tasks, where the testing relations are previously unseen in pre-training or training. We choose several state-of-the-art FSRL methods, as follows: \textbf{GMatching} \cite{xiong2018one} uses a neighbor encoder and a matching network, assuming that all neighbors contribute equally. \textbf{FSKGC} \cite{zhang2020few} encodes neighbors with a fixed attention mechanism, and applies a recurrent autoencoder to aggregate the few-shot instances in the support set. \textbf{GANA} \cite{niu2021relational} improves on FSKGC by having a gated and attentive neighbor aggregator to capture valuable contextual semantics of each few-shot relation. \textbf{FAAN} \cite{sheng2020adaptive} introduces an adaptive neighbor encoder for different relation tasks. \textbf{HiRe} \cite{wu2023hierarchical} brings in a hierarchical relational learning framework which considers triplet-level contextual information in contrastive learning. \textbf{MetaR} \cite{chen2019meta} utilizes a MAML-based framework, which aims to learn a good initialization for the unseen relations, followed by an optimization-based adaptation.

\section{Implementation details}
For WIKI and FB15K-237, we directly use the pre-trained embeddings provided in \cite{xiong2018one,wang2021reform}. For UMLS, we obtain the pre-trained embedding using the popular TransE-pytorch implementation by Mklimasz\footnote{https://github.com/mklimasz/TransE-PyTorch}. Throughout all the experiments, the embedding dimension is set to 100 for FB15K-237 and UMLS, and 50 for WIKI. Where applicable, the maximum number of neighbors of one given entity is set to 50. All results reported are on the candidate set after removing relations with less than 10 candidates. For each model, some settings are tuned using the validation set, while the others follow their respective original papers. More details on the hyperparameter settings can be found in Table~\ref{hyperparameters}.

We train \model\ for 100,000 epochs, and select the most optimal model based on the validation  relations every 1,000 epochs with early stopping for a patience setting of 30. The mini-batch gradient descent is applied with batch size set as 64 for FB15K-237 and UMLS, and 128 for WIKI. The number of hidden neurons is set as 50 for all datasets. We use Adam \cite{kingma2014adam} with the initial learning rate of 0.001 to update parameters. The intensity of gradient update is fixed at 5. The number of positive and negative triplets in each query set is 3 in FB15K-237 and UMLS, and 10 in WIKI. All experiments are conducted on an RTX3090 GPU server in Linux.

\section{Number of hops in $\mathcal{N}_{e}$}\label{hops}
\begin{table}[t]
\centering
\caption{Impact of the number of hops in neighborhood contexts $\mathcal{N}_e$ in the 3-shot setting.}
\label{layers} 
\small
\addtolength{\tabcolsep}{-0.6mm}
%\resizebox{1.0\textwidth}{!}{%
\begin{tabu}{@{}c|c|cc@{}} 
  \toprule
  Dataset &  No. of hops & MRR & Hit@10   \\
  \midrule
 \multirow{4}{*}{Wiki} & 1-hop  &0.347$\pm$.006 &0.454$\pm$.012 \\
 & 2-hop   &    0.353$\pm$.009 & 0.457$\pm$.010  \\
 & 3-hop  &    0.321$\pm$.005 & 0.415$\pm$.009 \\
\midrule
 \multirow{4}{*}{FB15K-237} & 1-hop  &0.405$\pm$.012 &0.575$\pm$.014 \\
 & 2-hop   &    0.402$\pm$.010 & 0.568$\pm$.016  \\
 & 3-hop  &    0.379$\pm$.014 & 0.531$\pm$.014 \\
\midrule
 \multirow{4}{*}{UMLS} & 1-hop  &0.608$\pm$.067 &0.780$\pm$.044 \\
 & 2-hop   &    0.543$\pm$.051 & 0.697$\pm$.049  \\
 & 3-hop  &    0.454$\pm$.048 & 0.612$\pm$.046 \\
 
\bottomrule
\end{tabu}
%}
\end{table}

As observed in Table~\ref{layers}, increasing the number of hops considered in neighborhood contexts $\mathcal{N}_e$ for data augmentation generally lead to performance degradation. For instance, on the UMLS dataset, the performance at the 3-hop setting is reduced by 25.3\% in MRR  compared to the original 1-hop setting in \model. This could be due to the additional noise introduced when considering a broader neighborhood, which can adversely affect model performance, especially for smaller datasets like UMLS, as they are more vulnerable to noise during model training.

\end{document}